\documentclass{article}

\usepackage{microtype}
\usepackage{graphicx}
\usepackage{subfigure}
\usepackage{booktabs} %

\usepackage{amsmath,bm}
\usepackage{bbm}
\usepackage{booktabs}
\usepackage{float}
\usepackage{color}
\usepackage{xcolor}
\usepackage{algorithm,algorithmic}
\usepackage{amsfonts}

\DeclareMathOperator*{\argmax}{arg\,max}

\usepackage{hyperref}

\usepackage[accepted]{icml2019}

\icmltitlerunning{Adaptive Model Selection for Airline Pricing}

\begin{document}

\twocolumn[
\icmltitle{Adaptive Model Selection Framework: An Application to Airline Pricing}

\icmlsetsymbol{equal}{*}

\begin{icmlauthorlist}
\icmlauthor{Naman Shukla}{equal,uiuc,deep}
\icmlauthor{Arinbj\"orn Kolbeinsson}{equal,ic}
\icmlauthor{Lavanya Marla}{uiuc}
\icmlauthor{Kartik Yellepeddi}{deep}
\end{icmlauthorlist}

\icmlaffiliation{uiuc}{Industrial and Enterprise Systems Engineering
University of Illinois, Urbana-Champaign, USA}
\icmlaffiliation{deep}{Deepair Solutions, London, UK}
\icmlaffiliation{ic}{Imperial College London, London, UK}

\icmlcorrespondingauthor{Naman Shukla}{naman@deepair.io}
\icmlcorrespondingauthor{Kartik Yellepeddi}{kartik@deepair.io}

\icmlkeywords{Machine Learning, ICML}

\vskip 0.3in
]

\printAffiliationsAndNotice{\icmlEqualContribution} %

\begin{abstract}
Multiple machine learning and prediction models are often used for the same prediction or recommendation task. In our recent work, where we develop and deploy airline ancillary pricing models in an online setting, we found that among multiple pricing models developed, no one model clearly dominates other models for all incoming customer requests. Thus, as algorithm designers, we face an exploration - exploitation dilemma. In this work, we introduce an adaptive meta-decision framework that uses Thompson sampling, a popular multi-armed bandit solution method, to route customer requests to various pricing models based on their online performance. We show that this adaptive approach outperform a uniformly random selection policy by improving the expected revenue per offer by 43\% and conversion score by 58\% in an offline simulation. 
\end{abstract}
\vspace{0.4cm}
\section{Introduction}
In recent years, ancillaries such as bags, meals, wifi service and extra leg room have become a major source of revenue and profitability for airlines \cite{IdeaWorks, bockelie2017incorporating}. Conventional pricing strategies based on static business rules do not respond to changing market conditions and do not match customer willingness to pay, leaving potential revenue sources untapped. In our recent work \cite{shukla2019dynamic}, we demonstrate that machine learning approaches that dynamically price such ancillaries based on customer context lead to higher revenue on average for the airline compared to conventional pricing schemes. Specifically, we presented three different airline ancillary pricing models that provided dynamic price recommendations specific to each customer interaction and improved the revenue per offer for the airline. We also measured the real-world business impact of these approaches by deploying them in an A/B test on an airline's internet booking website. As a part of that research, in order to measure the relative performance of these models they were all deployed in parallel on the airline's booking system. The online business performance of our deployed models was better than human on an average, but no single model outperformed the other deployed models in all customer contexts. Moreover, we were also developing new models that were performing well in the offline setting and showed the promise to do better in an online setting. As a result, we face an exploration-exploitation dilemma -- do we exploit (a) the one single model that does best on an average,  or (b) explore other models that do better in a different context or (c) utilize offline metrics? Therefore, we present a meta-decision framework that addresses this issue.

Our contributions in this work are as follows.
\vspace{-0.35cm}
\begin{itemize}
\itemsep-0.05cm
    \item We develop an approach that uses multi-armed bandit method to actively and adaptively route pricing requests to multiple models to further improve revenue.
    \item We test this approach in a rigorously constructed simulation environment to demonstrate that a improved routing scheme that improves business metrics can be achieved.
    \item We lay a foundation for future research to use contextual multi-armed bandit methods and perform online testing of this approach.
\end{itemize}

\subsection{Related work}

Ensemble methods are meta-learning algorithms that combine multiple machine learning methods to reduce error and variance \cite{kotsiantis2007supervised}. Several ensemble learning methods have been proposed to improve personalization systems. Meta-recommenders were pioneered by \citet{schafer2002meta} in their work on combining multiple information sources and recommendation techniques for improved recommendations. More recent advancements in ensemble learning in the context of the 'bucket of models' framework include model selection using cross-validation techniques \cite{arlot2010survey}. Most related to our work is the use of contextual multi-armed bandits for recommendation in an online setting \cite{zeng2016online}. The notable difference to our approach in this paper is that they apply the bandit directly to the problem; whereas we first generate recommendations using an assortment of diverse models. Multi-armed bandits have also been used to control for false-discovery rate in sequential A/B tests \cite{yang2017framework}. However, that method does not allow for maximizing an arbitrary outcome (such as revenue or offer conversion score), which is a necessary property for our required solution. 

\vspace{-0.25cm}
\section{Our Approach}
Among a set of models that price ancillaries, our objective is to direct online traffic (customers) more effectively to better performing models that predict customer willingness to pay more accurately to maximize revenue. To formalize this, we model the online traffic share directed to each pricing model as a tunable meta-parameter. Learning this parameter is therefore critical to identify the ``winning'' model and separate it from other models. Based on customers' responses (to purchase an ancillary or not), we can reinforce the meta-parameters to adapt to the outcome. We model this problem from a reinforcement learning perspective as a sequence of actions (selecting a model) and rewards (offer accepted or not) that can be learned using a multi-armed bandit.

\subsection{The Multi-Armed Bandit approach}

 Solving the multi-armed bandit problem involves choosing an action (pulling an 'arm'), from a given action space, when the reward associated with each action is only partially known \cite{auer2002finite, sutton2018reinforcement}. The environment reveals a reward after an action is taken, thereby giving information about both the environment and the action's reward. The objective of the problem is to maximize the expected reward or minimize the cumulative expected regret \cite{bubeck2012regret} from a series of actions. The bandit problem involves a trade off between exploration and exploitation since the reward is only revealed for the chosen action. In our framework, each of the deployed ancillary pricing models is viewed as an arm. This enables the decision-maker to direct online traffic based on the customer's response to the price offered by that arm. Hence, this meta-learning approach provides a trainable mechanism to allocate customer traffic to the arms that provide best conversion scores. We use the Thompson sampling algorithm, a popular multi-armed bandit solution approach, which allows us to exploit the current winning arm and explore seemingly inferior arms, or newly introduced arms that might outperform the current best arm \cite{chapelle2011empirical, russo2018tutorial}.

\subsection{Thompson sampling for the Bernoulli Bandit}
In a Bernoulli bandit problem, there are a total of $|A|$ valid actions, where an action $a_{t} \in A$ at time $t$ produces a reward $r_{t} \in \{0, 1\}$ of one with probability $\theta_{a}$ and zero with probability $1 - \theta_{a}$. The mean reward $\theta = (\theta_{1}, \dots ,\theta_{|A|})$ is assumed to be unknown, but is stationary over time. The agent begins with an independent prior belief over each $\theta_a, a = 1,\dots,|A|$. As observations are gathered, the distribution is updated according to Bayes' rule. As described in equation \eqref{eq:beta}, the priors are assumed to be beta-distributed with parameters $\alpha_a \in \{\alpha_1, \dots , \alpha_{|A|}\}$ and $\beta_a \in \{\beta_1, \dots , \beta_{|A|}\}$. In particular, the exact prior probability density function given for an action $a$ is
\begin{equation}
\label{eq:beta}
p_{beta}(\theta_a) = \frac{\Gamma(\alpha_a + \beta_a)}{\Gamma(\alpha_a)\Gamma(\beta_a)} \theta_{a}^{(\alpha_a - 1)}(1-\theta_{a})^{(\beta_a - 1)},
\end{equation}
where $\Gamma$ denotes the gamma function. The beta distribution is particularly suited to this computation because it is a conjugate prior to the Bernoulli distribution.

Algorithm \ref{alg:ts} describes the method used for choosing models within our simulation environment (Figure \ref{fig:flowchart}, Step 2). $A$ is the set of models (described in Section \ref{Online implementation}). $x_t$ is the model chosen to provide the pricing recommendation, and the corresponding reward $r_t$ is 1 if the ancillary is purchased and 0 otherwise. 

\begin{algorithm}
\caption{Thompson sampling for the Bernoulli Bandit}\label{algo:TS}
\label{alg:ts}
  \begin{algorithmic}[1]
    \FOR{$t = 1, 2, \dots$}
      \FOR{$a = 1, \dots, |A|$} %
        \STATE Sample $\hat{\theta_a} \sim \text{Beta}(\alpha_a, \beta_a)$
      \ENDFOR
      \STATE $x_t \leftarrow$ $\argmax_a \hat{\theta_a}$ \ \ \ \ %
      \STATE Apply $x_t$ \ \ \ \ \ \ \ \ \ \ \ \ \ \ %
       \STATE Observe $y_t$, $r_t$ \ \ \ \ \ \ \ \ \ \ \ \ %
      \STATE $(\alpha_{x_t}, \beta_{x_t}) \leftarrow (\alpha_{x_t} + r_t, \beta_{x_t} + 1 - r_t)$ \ \ \ %
    \ENDFOR
  \end{algorithmic} 
\end{algorithm}

\vspace{-0.25cm}
\section{Results}
\subsection{Simulations with synthetic data}
In this section we discuss an offline simulation performed on synthetic data to show the convergence score of Thompson sampling on the ancillary pricing models developed in our work. Specifically, we create customer session samples where the three experimental models have a success rate in capturing customer willingness to pay of $0.45$, $0.55$ and $0.60$, respectively; and then aim to infer these probabilities through Thompson sampling from the caller streams.

We start the simulation with the prior $\theta^{0} = Beta(\alpha=1, \beta=1)$, which corresponds to a uniform prior between 0 and 1. The run is then simulated for 2,000 steps and target probabilities are recorded. Table \ref{tab:simulation} demonstrates that Thompson sampling converges to the latent probability distribution (the success rates $\theta$ of each model) within a reasonable number of iterations, indicating its potential for deployment in real-world pricing settings. 
\vspace{-0.25cm}

\begin{table}[h]
\centering
\caption{Simulation results for Thompson sampling after 2,000 steps.}

\begin{tabular}{@{}cccc@{}}
\toprule
$\theta$ & True probability & Inferred probability & \# of trials \\ \midrule
$\theta_1$    & 0.45      & 0.45           & 70     \\
$\theta_2$    & 0.55      & 0.55           & 305    \\
$\theta_3$    & 0.60      & 0.61           & 1625   \\ \bottomrule
\end{tabular}
\label{tab:simulation}
\end{table}

\subsection{Pricing under simulation environment }
\label{Online implementation}

\begin{figure}[h]
\centering
\includegraphics[scale=0.5]{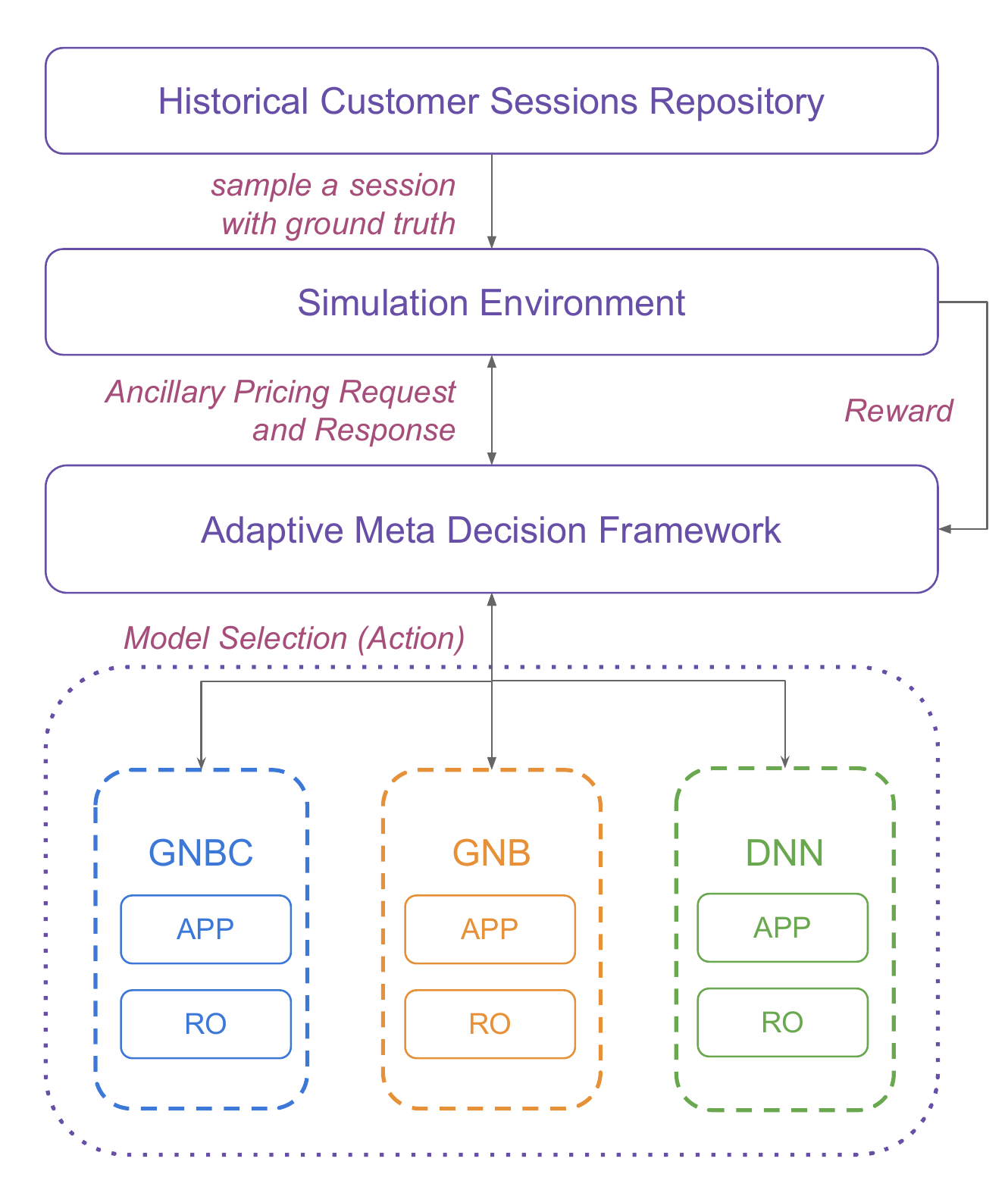}
\caption{Offline testing simulation environment setup}
\label{fig:flowchart}
\end{figure}

\begin{figure}[h]
\centering
\includegraphics[scale=0.5]{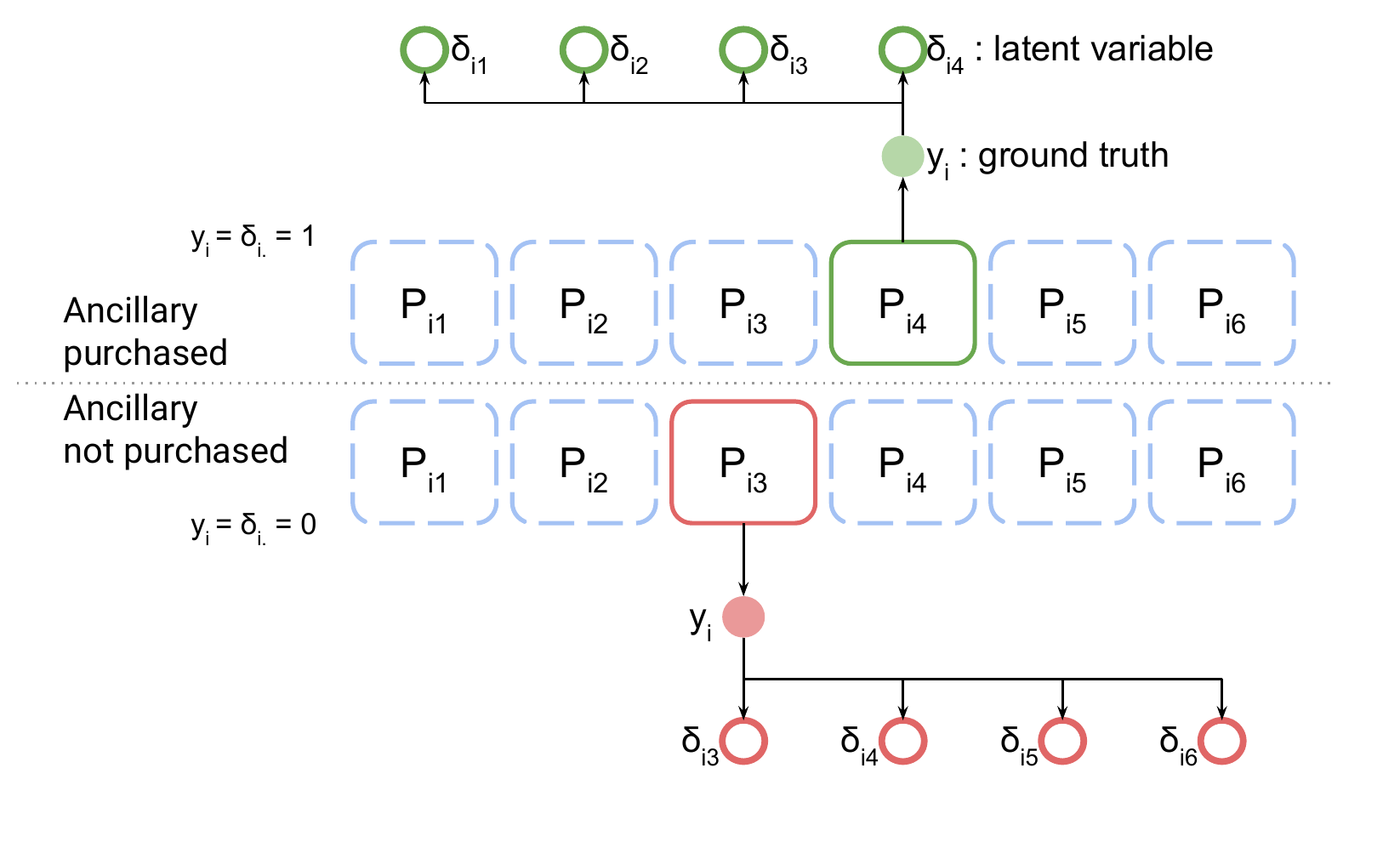}
\caption{Latent variable $\delta$ mapping from ground truth $y$ where prices are arranged in ascending order from left to right}
\label{fig:delta}
\end{figure}

We now apply the Thompson sampling algorithm to a testing environment constructed based on real-world data. Figure \ref{fig:flowchart} shows the overview of our testing setup for evaluating the adaptive meta decision framework. Our simulation environment models customer behavior by sampling sessions based on 6 months of historically collected data, amounting to a total of about 16000 sessions. In our models, we make monotonicity assumptions for consistency. First, we assume that if a customer is willing to purchase a product for price $p$, they are willing to purchase the same product at a price $p' < p$. Similarly, if a customer is unwilling to purchase a product at price $p$, they will be unwilling to purchase at a price $p' > p$. We define $\delta_{ij}$, shown in Figure \ref{fig:delta}, as a latent variable that ensures this assumption by taking the historical ground truth $y_{i}$ (purchased by customer or not) into account. Hence, the latent variable can be used as the reward response $r_{i}$ from simulator for the offered price, given that customer session.

We consider the following three pricing models, presented in \cite{shukla2019dynamic}, in which pricing is modeled as a 2-step processes. First, the purchase probability of the ancillary is estimated as a function of the price; and second, a price that optimizes revenue is offered to the customer. Further details on these models can be found in our previous work \cite{shukla2019dynamic}. 

\vspace{-0.25cm}
\begin{enumerate}
\item \textbf{Gaussian Naive Bayes (GNB)}: This two-stage pricing model uses a Gaussian Naive Bayes (GNB) model for ancillary purchase probability prediction and a pre-calibrated logistic price mapping function for revenue optimization.
\item \textbf{Gaussian Naive Bayes with clustered features (GNBC)}: This two-stage pricing model uses a Gaussian Naive Bayes with clustered features (GNBC) model for ancillary purchase probability prediction and a pre-calibrated logistic price mapping function for revenue optimization.
\item  \textbf{Deep-Neural Network (DNN)}: This two-stage pricing model uses a Deep-Neural Network (DNN) trained using a weighted cross-entropy loss function for ancillary purchase probability estimation. For price optimization, we implement a simple discrete exhaustive search algorithm that finds the optimal price point within the permissible pricing range.
\end{enumerate}
\vspace{-0.1cm}

\begin{figure}[h]
\centering
\includegraphics[scale=0.18]{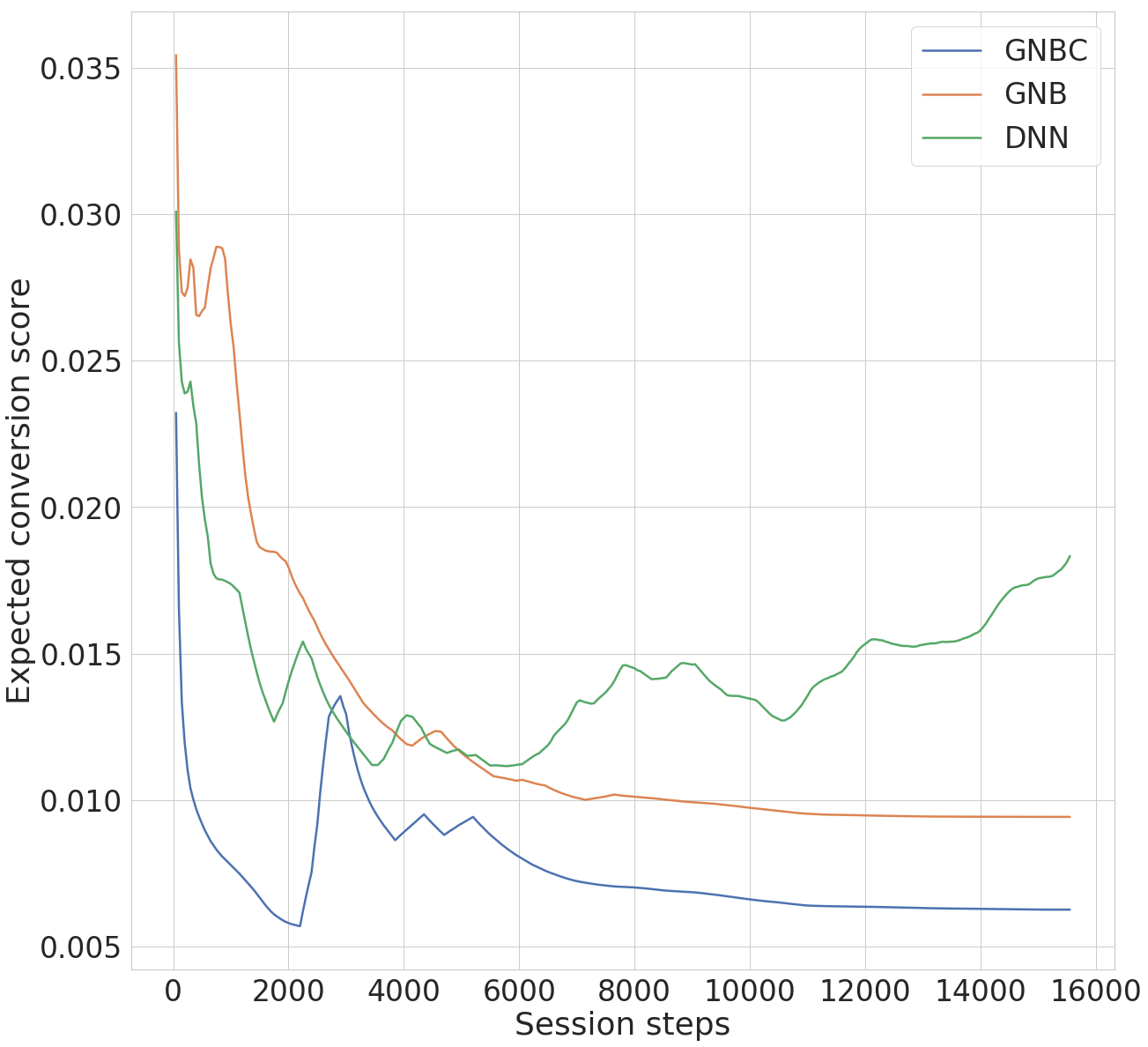}
\caption{Conversion score of each arm using Thompson sampling}
\label{fig:successrate}
\end{figure}

\begin{figure}[h]
\centering
\includegraphics[scale=0.18]{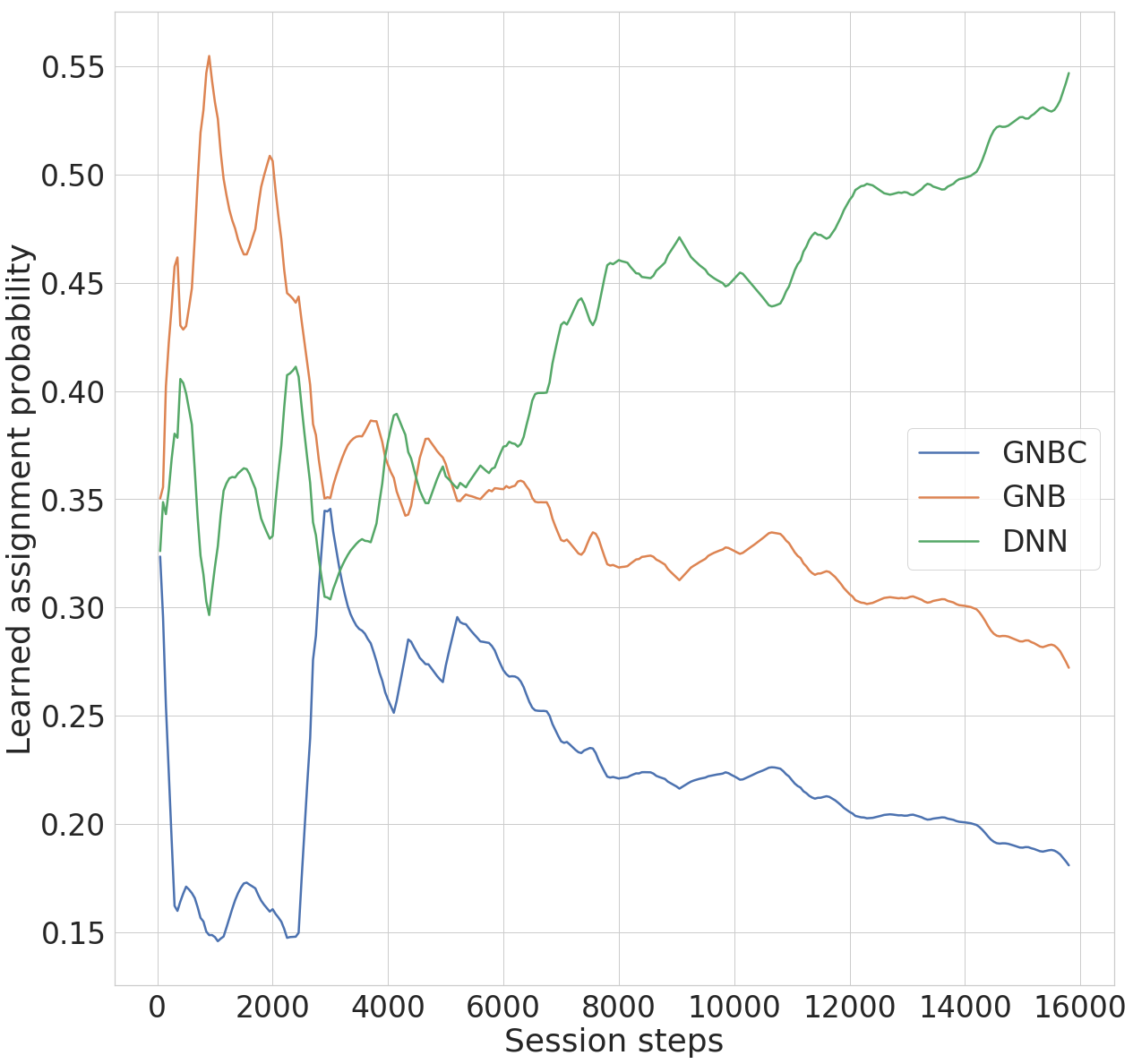}
\caption{Learned Assignment Probability}
\label{fig:assprob}
\end{figure}

Our simulation results using Thompson sampling as a meta learning framework are shown in Table \ref{tab:pricing}. The expected offer conversion scores for each model, where each model is considered as an arm of the multi-armed bandit, is shown in Figure \ref{fig:successrate}. Also, Figure \ref{fig:assprob} shows the evolution of the assignment probability learned using Thompson sampling. Figure \ref{fig:successrate} shows that the probability of success for GNB is higher than the others for the first 5,000 sessions. Subsequently, it is apparent that the DNN model dominates as the sessions progress. Consequently, learned assignment probability with the number of sessions (steps) for DNN shows an increasing trend, whereas the learned assignment probability for GNB and GNBC shows a decreasing trend, with both breaking even around 5,000 steps. We observe DNN significantly outperforms GNB and GNBC. This is reflected in the assignment probability that is learned using Thompson sampling. The reason behind DNN outperforming other models is that the recommended prices by DNN is comparatively lower than the other two, on average. Since the reward is given only to affirmative response using simulator which is operating on willingness to pay assumption, only lower prices with ground truth $y_i$ equal to $1$ are awarded. Provided that the simulator is biased towards the model which prices conservatively rather than aggressively, the routing probability is able to learn this trend through the meta learning based on Thompson sampling.

\begin{table}[]
\centering
\caption{Thompson sampling converged values for ancillary pricing using simulated environment. The offer conversion score is the percent of offers generated that were accepted by customers.}
\begin{tabular}{@{}cccc@{}}
\toprule
Model & Assignment probability & Offer conversion score \\ \midrule
GNBC    & 0.177    & 0.627\%     \\
GNB    & 0.266    & 0.943\%     \\
\textbf{DNN}    & \textbf{0.557}    & \textbf{1.976\%}     \\ \bottomrule
\end{tabular}
\label{tab:pricing}
\end{table}

\begin{table}[h]
\centering
\caption{Revenue per offer and Offer conversion scores for different models } 
\begin{tabular}{@{}lcc@{}}
\toprule
Models    & Revenue per offer & Conversion score \\ \midrule
Only GNBC & 0.57             & 0.61\%             \\
Only GNB  & 0.84             & 0.93\%             \\
Only DNN  & 1.41              & 1.71\%             \\
Random    & 0.93             & 1.00\%             \\
MAB       & 1.33             & 1.58\%             \\ \bottomrule
\end{tabular}
\label{tab:rpo}
\end{table}
\vspace{-0.25cm}

\begin{figure}[h]
\centering
\includegraphics[scale=0.18]{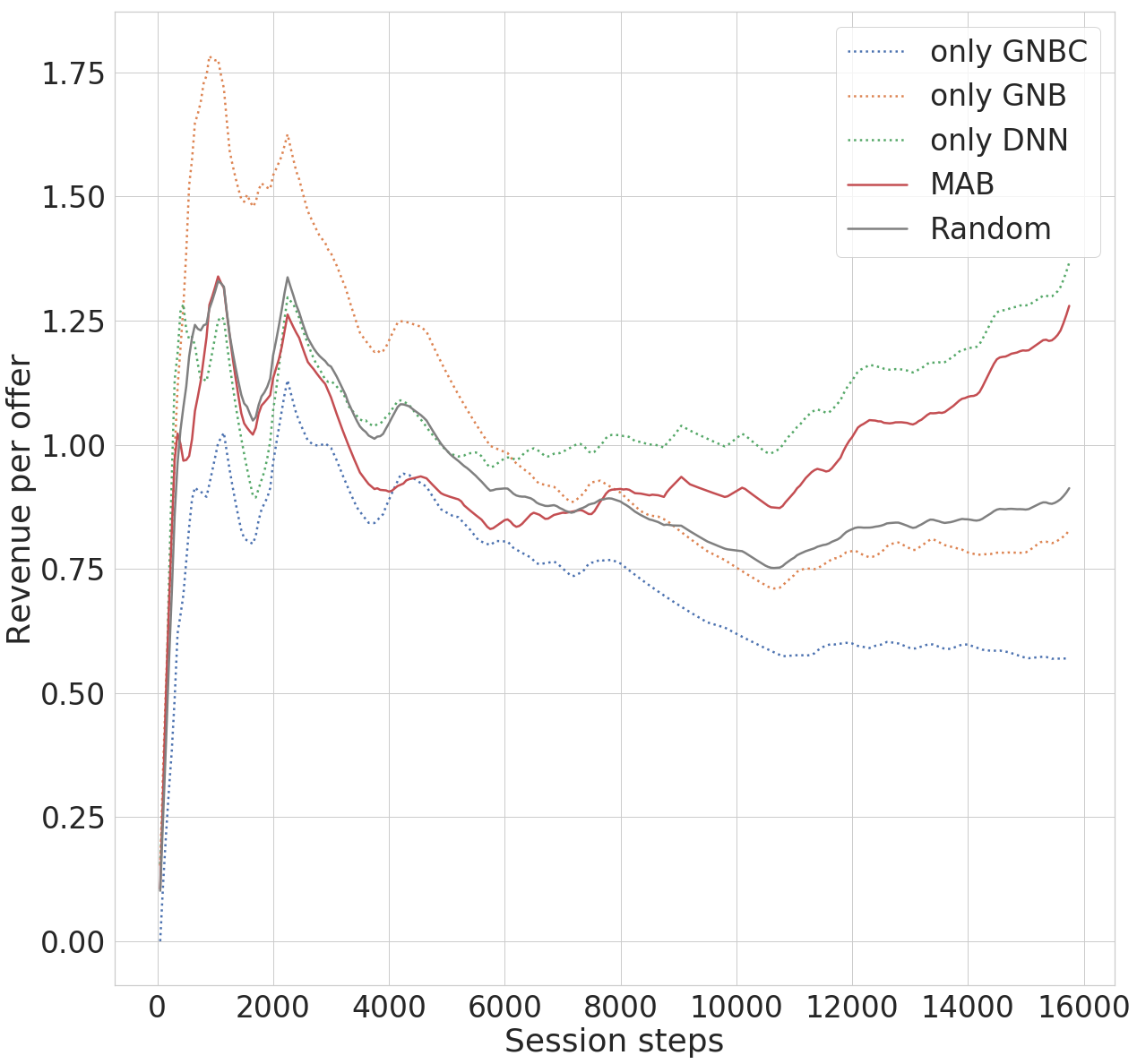}
\caption{Expected revenue per offer}
\label{fig:rpo}
\end{figure}

From a business perspective, revenue per offer is another crucial metric. We measure expected revenue per offer in the following scenarios :multi-armed bandit (MAB) as an adaptive meta-decision framework, random selection of pricing models with equal probability, and each individual model (GNBC, GNB or DNN) recommending ancillary price. Figure \ref{fig:rpo} shows expected revenue per offer in the simulation environment for the six month, 16000 sessions period. Figure \ref{fig:rpo} shows that the revenue per offer generated by only GNB is highest until 6,000 sessions. After 6,000 sessions, DNN generates better revenue per offer. Hence, an ideal meta-decision framework should select the model that generates the highest revenue for each offer, i.e., GNB until 6,000 sessions and DNN afterwards. Since information about the customer's response is unavailable prior to the model selection, the model that will generate the highest revenue per offer is not deterministic. Hence, exploration is used in the MAB-based approach. The trend for the MAB using Thompson sampling (red line) is similar to the ideal meta-decision framework. Table \ref{tab:rpo} shows the expected revenue per offer and conversion scores at the end of the simulation. The MAB metamodel generates 43\% more revenue per offer and 58\% more conversion than random selection. However, the cost of exploration is reflected in the revenue per offer because the DNN is able to get 6\% higher revenue per offer as well as 8\% higher conversion score than the MAB by the end of the simulation.

\vspace{-0.25cm}
\section{Discussion}
The objective of Thompson sampling is to maximize the conversion score. Nonetheless, the improvement to the revenue per offer is entirely implicit but not guaranteed. In the future, we plan to adjust the formulation of the multi-armed bandit method to maximize for revenue per offer directly. Another important benefit of the Thompson sampling based approach is active protection against revenue loss from directing traffic to sub-optimal models. This meta-decision making framework allows for instant and automated adjustments to the traffic assignment ratios. For example, new models that performs badly compared to previously deployed models will get iteratively fewer sessions assigned to it, reducing the amount of lost revenue.

\vspace{-0.15cm}
\subsection{Constraints, Context and Caution} \label{CCC}
Another extension of the meta-decision framework based on the multi-armed bandit approach addresses contextual online decision problems \cite{russo2018tutorial}. In these problems, the choice of the arm $a_t$ in a multi-armed bandit also depends on an independent random variable $z_t$ that the agent observes prior to making the decision. In such scenarios, the conditional distribution of the response $y_t$ is of the form $p_{\theta}(\cdot |a_t, z_t)$. Contextual bandit problems of this kind can be addressed through augmenting the action space and introducing time-varying constraint sets by viewing action and constraint together as $\tilde{a_t} = (a_t, z_t)$, with each arm (choice) represented as $\tilde{A_t} = \{(a_t,z_t : a\in A)\}$, where $A$ is the set from which $x_t$ must be chosen. 

In settings where models have a particular baseline criteria to maintain, caution-based sampling can be used \cite{chapelle2011empirical, russo2018tutorial}. This can be accomplished through constraining actions for each time $t^{th}$ step to have lower bound on expected average reward as $A_t = \{ a \in A : \mathbb{E}[r_t| a_t = a] \geq \tilde{r}\}$. This ensures that expected average reward at least exceeds $\tilde{r}$ using such actions. 

\vspace{-0.15cm}
\subsection{Non-stationary systems and Concurrence} \label{Non stationary systems and Concurrence}
So far, we discussed settings in which target meta-parameters are constant over time i.e. belong to a stationary system. In practice, dynamic pricing problems are time dependent and thereby non-stationary; and are more appropriately modeled by time-varying parameters, such that reward is generated by $p_{\theta_t}(\cdot |a_t)$. In such contexts, the multi-armed bandit approach will never stop exploring the arms, which could be a potential drawback. A more robust method involves ignoring all historical observations made prior to a certain time period $\tau$ in the past \cite{cortes2017multi}. Decision-makers can produce a posterior distribution after every time step $t$ based on the prior and condition only on the most recent $\tau$ actions and observations. Model parameters are sampled from this distribution, and an action is selected to optimize the associated model. 

An alternative approach is to view dynamic pricing recommendations as a weighted sum of prices from multiple arms, a concept referred to as concurrence. In this case, the decision-maker takes multiple actions (arms) concurrently. Concurrency can be predefined with number of fixed arms to be pulled every time or it could be coupled with baseline caution (discussed in Section \ref{CCC}). This approach is similar to an approach based on an ensemble of models.

\section{Conclusion}
We are able to successfully demonstrate, for a dynamic pricing problem in an offline setting, that a multi-armed bandit approach can be used to adaptively learn a better routing scheme to generate 43\% more revenue per offer and 58\% more conversion than a randomly selected model, when dealing with a bucket of models. However, this approach takes a slight toll on revenue per offer and conversion score due to the exploration phase, in comparison to the best performing model alone. We are currently working on deploying this in an online setting, where purchasing customers reveal the ground truth in real-time. In the future, we plan to extend our model to contextual multi-armed bandits with caution and concurrence, to optimize for different business metrics. We will also model customers' elasticity towards an offered ancillary to enable the simulator environment to model ground truth more closely. Finally, although this paper focuses specifically on ancillary pricing for airlines, the framework presented here can be applied to any machine learning problem where a bucket of online models and a pipeline of new models are waiting to be deployed.

\section*{Acknowledgements}
We sincerely acknowledge our airline partners for their continuing support. The academic partners are also thankful to deepair (www.deepair.io) for funding this research.

\bibliography{refs}
\bibliographystyle{icml2019}

\end{document}